\documentclass{article}

\usepackage{microtype}
\usepackage{graphicx}
\usepackage{subcaption}
\usepackage{booktabs} 
\usepackage{multirow} 
\usepackage{hyperref}
\usepackage{enumitem}
\usepackage{tcolorbox}
\usepackage{colortbl}
\usepackage{multirow}
\definecolor{findingborder}{RGB}{0, 85, 80} 
\definecolor{findingbg}{RGB}{235, 235, 235} 

\usepackage[accepted]{icml2026}

\usepackage{amsmath}
\usepackage{amssymb}
\usepackage{mathtools}
\usepackage{amsthm}
\usepackage{algorithm}
\usepackage{algorithmic}

\usepackage[capitalize,noabbrev]{cleveref}

\theoremstyle{plain}
\newtheorem{theorem}{Theorem}[section]

\theoremstyle{definition}
\newtheorem{definition}[theorem]{Definition}

\theoremstyle{remark}

\newcommand{\argmin}{\operatorname{argmin}}
\newcommand{\softmax}{\operatorname{softmax}}
\newcommand{\topk}{\operatorname{top-k}}
\newcommand{\sg}{\operatorname{sg}}

\newcommand{\name}{DOT-MoE}
\usepackage[textsize=tiny]{todonotes}

\icmltitlerunning{DOT-MoE: Differentiable Optimal Transport for MoEfication}

\begin{document}

\twocolumn[
  \icmltitle{\name: Differentiable Optimal Transport for MoEfication}



  \icmlsetsymbol{equal}{*}

  \begin{icmlauthorlist}
    \icmlauthor{Udbhav Bamba}{equal,amz}
    \icmlauthor{Arnav Chavan}{equal,amz}
    \icmlauthor{Aryamaan Thakur}{equal,amz}
    \icmlauthor{Steve Teig}{amz}
    \icmlauthor{Deepak Gupta}{amz}

  \end{icmlauthorlist}

  \icmlaffiliation{amz}{Amazon}
  \icmlcorrespondingauthor{Udbhav Bamba}{udbhb@amazon.com}

  \icmlkeywords{Machine Learning, ICML}

  \vskip 0.3in
]



\printAffiliationsAndNotice{\icmlEqualContribution}

\begin{abstract}
The scaling of Large Language Models (LLMs) has driven significant performance gains but created substantial challenges in inference efficiency. While Mixture of Experts (MoEs) architectures address this by decoupling model size from inference cost, training MoEs from scratch is often unstable and compute intensive. Conversion of pre-trained dense models into sparse MoEs has emerged as an alternative solution; however, existing methods typically rely on heuristic neuron clustering or random splitting to partition the Feed-Forward Network (FFN) into experts. In this work, we propose DOT-MoE, a novel framework that formulates the decomposition of dense layers as a Differentiable Optimal Transport (DOT) problem. Instead of static heuristics, we model neuron assignment as a balanced transport problem, utilizing differentiable Sinkhorn-Knopp iterations to enforce strict expert capacity constraints. Furthermore, we utilize Straight-Through Estimators (STE) to jointly learn the discrete neuron-to-expert assignment and the token-to-expert routing policy end-to-end. Extensive experiments across multiple architectures and benchmarks demonstrate that DOT-MoE significantly outperforms structured pruning, heuristic clustering, and random-split baselines, retaining 90\% of the original dense model's performance while reducing active parameters by 50\%.
\end{abstract}
\section{Introduction}
The rapid scaling of Large Language Models (LLMs) has led to remarkable capabilities in natural language understanding and generation. However, this performance comes at a prohibitive computational cost. As model dimensions grow, the dense activation patterns of standard Transformers \cite{vaswani2017attention}, where every parameter is active for every input token, result in unsustainable inference latency and resource consumption. To address this efficiency bottleneck, Mixture-of-Experts (MoE) \cite{shazeer2017sparsely, lepikhin2020gshard, jiang2024mixtral, fedus2022switch} architectures have emerged as a promising solution. By routing tokens to a small subset of \textit{expert} sub-networks, MoEs decouple model size from inference cost. For example, the recent Qwen3-30B-A3B \cite{yang2025qwen3technicalreport} MoE architecture comprises a total of 30.5B parameters in the network, however, only 3.3B parameters are activated per token during inference.

Despite their inference efficiency, training MoE models from scratch is notoriously data-hungry and unstable, often requiring complex load-balancing auxiliaries \cite{zoph2022st, fedus2022switch}. Consequently, a new paradigm has gained traction: MoEfication \cite{zhang-etal-2022-moefication}, or the conversion of pre-trained dense models into sparse MoEs. This approach leverages the high-quality representations of existing dense checkpoints, transforming the Feed-Forward Network (FFN) in each block into sparse experts to reduce inference FLOPs without the cost of pre-training from scratch. Conceptually, this process can be viewed as a form of dynamic structural pruning. Unlike static pruning methods~\citep{frantar2023sparsegpt, sunsimple, ashkboos2024slicegpt, gao2024disp}, which permanently remove parameters and often degrade performance by erasing \textit{long-tail} knowledge essential for LLMs \cite{lele2025rethinking}, MoEfication retains the full parameter set but activates it selectively. By dynamically pruning the network structure conditioned on the input, it maintains the high capacity of the dense model while achieving the efficiency of sparse execution.

The core challenge in this conversion is neuron assignment: determining how to partition the thousands of intermediate neurons in a dense FFN into discrete, independent, and functionally coherent experts. Existing approaches to dense-to-MoE conversion \cite{llamamoe, llamamoev2, cmoe} rely largely on heuristic strategies for neuron assignment. While effective, these methods often treat neuron assignment and router training as separate processes. They lack a unified, differentiable framework that guarantees balanced expert capacity while simultaneously optimizing for the semantic routing of tokens.

In this work, we propose DOT-MoE, a novel framework that formulates the decomposition of dense layers into experts as a Differentiable Optimal Transport (DOT) problem. We enable end-to-end learning of both the expert decomposition and the routing mechanism by using differentiable Sinkhorn-Knopp iterations \cite{knight2008sinkhorn, sinkhorn1967concerning}. Unlike existing methods \cite{cmoe, llamamoe, llamamoev2} which freeze the assignment and then train the router, DOT-MoE allows the router and the expert assignment to co-adapt. Instead of relying on static heuristics, we view neuron assignment as a balanced transport problem where neurons must be \textit{transported} to experts under strict capacity constraints. This is achieved with Straight-Through Estimators (STEs) \cite{bengio2013estimating} that allow gradients to flow through the discrete assignment decisions. This ensures that experts are not just random collections of neurons, but balanced functional units optimized specifically for the routing policy.

Our main contributions are:
\begin{enumerate}[noitemsep]
\item We introduce an optimal transport framework for dense-to-MoE conversion, formulating neuron assignment as a balanced transport problem with differentiable Sinkhorn iterations.
\item We develop a dual-level assignment mechanism that jointly optimizes neuron-to-expert decomposition and token-to-expert routing via complementary straight-through estimators, enabling the router and expert structure to co-adapt unlike prior methods that treat these as separate stages.
\item We demonstrate through extensive experiments across three model families (LLaMA-2, LLaMA-3, Qwen2.5) and six benchmarks that DOT-MoE outperforms both structured pruning and existing MoEfication methods, retaining 90\% of the original dense model's performance at 50\% parametric count.
\end{enumerate}

\section{Background and Motivation}
\label{sec:background}

\subsection{Preliminaries: FFN Layers in Transformers}
The FFN constitutes the majority of parameters in transformer-based language models, typically accounting for approximately two-thirds of total model parameters. It processes hidden states through a two-stage projection:
\begin{align}
 \mathbf{H} &= \sigma(\mathbf{x}\mathbf{W}_{\text{gate}}) \odot (\mathbf{x}\mathbf{W}_{\text{up}}) \label{eq:ffn_intermediate}\\
 \text{FFN}(\mathbf{x}) &= \mathbf{H}\mathbf{W}_{\text{down}} \label{eq:ffn_output}
\end{align}
where $\mathbf{x} \in \mathbb{R}^{d}$ is the input hidden state, $\mathbf{W}_{\text{gate}}, \mathbf{W}_{\text{up}} \in \mathbb{R}^{d \times d_{\text{ffn}}}$ and $\mathbf{W}_{\text{down}} \in \mathbb{R}^{d_{\text{ffn}} \times d}$ are weight matrices, $\sigma(\cdot)$ is an activation function (e.g., SiLU), and $\odot$ denotes element-wise multiplication.

\subsection{Existing Approaches for Efficient Inference}

Reducing the computational cost of large language models while preserving their capabilities remains a central challenge. Existing approaches fall into two categories: \textit{structured pruning}, which permanently removes model components, and \textit{dense-to-MoE conversion}, which converts dense layers into sparse mixtures of experts.

\subsubsection{Structured Pruning}

Structured pruning improves efficiency by permanently removing structured components such as neurons, channels, attention heads, or entire layers from dense LLMs~\cite{wang2020structured}. Representative methods such as ShortGPT \cite{men2025shortgpt} and SliceGPT \cite{ashkboos2024slicegpt} exploit structural redundancies via layer- or subspace-level importance criteria, achieving hardware-friendly sparsity and predictable inference speedups. More recent approaches, notably DISP-LLM \cite{gao2024disp}, substantially improve the accuracy-efficiency trade-off by enabling flexible, dimension-wise structural pruning and consistently outperform prior structured pruning baselines. However, structured pruning \textit{irreversibly removes model capacity}, which often leads to sharp performance degradation, particularly at high compression ratios.

\subsubsection{Dense-to-MoE Conversion}

MoEfication~\citep{zhang-etal-2022-moefication} introduced the paradigm of converting dense FFN layers into sparse MoE layers, preserving total model capacity while reducing per-token computation. This is achieved by partitioning the $d_{\text{ffn}}$ intermediate neurons into $E$ disjoint expert groups, each containing $s = d_{\text{ffn}}/E$ neurons. A learned router $\mathbf{W}_r \in \mathbb{R}^{E \times d_{\text{model}}}$ selects the top-$k$ experts per token:
\begin{equation}
 \mathcal{I} = \text{top-}k\left(\text{softmax}(\mathbf{x}\mathbf{W}_r^\top), k\right)
 \label{eq:routing}
\end{equation}
This sparsification reduces computational cost from $\mathcal{O}(d_{\text{ffn}})$ to $\mathcal{O}(k \cdot d_{\text{ffn}}/E)$ per token. Existing dense-to-MoE methods differ primarily in how they assign neurons to experts: 

\textit{(i) Random Assignment. }LLaMA-MoE~\cite{llamamoe} randomly partitions neurons into experts and relies on extensive continued pre-training to recover performance. While simple, this approach requires substantial computational resources and provides no principled basis for expert specialization. 

\textit{(ii) Weight-based Clustering. }LTE and MoEfication~\citep{zhang-etal-2022-moefication, 2024Haizhonglte} clusters neurons based on the similarity of their projection weights $\mathbf{W}_{\text{gate}}$ and $\mathbf{W}_{\text{up}}$, assuming that neurons with similar input-side weights respond to similar input patterns. 

\textit{(iii) Activation-based Clustering. }LLaMA-MoE-v2~\cite{llamamoev2} assigns neurons to experts based on importance estimates derived from activations and gradients. CMoE~\cite{cmoe} clusters intermediate FFN activations $\mathbf{H}$ with balanced $k$-means, grouping neurons by empirical co-activation patterns.

\subsection{Limitation: Optimizing Proxies Instead of Outputs}

Existing approaches share a fundamental limitation: \textit{they optimize proxies for intermediate representations while neglecting the actual output}. Consider Equation~\ref{eq:ffn_output}: the output of the FFN depends on the interaction between intermediate activations $\mathbf{H}$ and down-projection weights $\mathbf{W}_{\text{down}}$. Structured pruning methods optimize layer-wise importance scores that ignore this interaction. Dense-to-MoE methods cluster based on input weights, intermediate activations, or co-activation patterns, all of which are proxies that fail to capture how each neuron ultimately contributes to the output.

To empirically validate this limitation, we conducted a controlled single-layer reconstruction analysis on LLaMA-2 and LLaMA-3 (see Appendix~\ref{app:motivation_experiment}). By isolating the expert construction strategy, we observe that methods relying on input-side statistics~\cite{llamamoe} or intermediate activations~\cite{llamamoev2, cmoe} incur mean squared errors ranging from $2\times$ to over $41\times$ higher than our proposed approach. These results confirm that preserving FFN fidelity requires explicitly modeling the neuron's contribution to the output, rather than optimizing based on proxies.

\section{Method}
\label{sec:method}

\subsection{Problem Formulation}

Our goal is to convert a dense FFN layer into a sparse MoE layer with $E$ experts, each containing $s$ intermediate neurons such that $s \cdot E = d_{\text{ffn}}$, while activating only $k < E$ experts per token. Crucially, we aim to achieve this conversion without full model fine-tuning, enabling efficient post-hoc sparsification of pretrained dense models.

The central challenge is neuron assignment: determining which 
of the $d_{\text{ffn}}$ 
neurons should be grouped into each expert. This problem is combinatorially intractable, with $\frac{d_{\text{ffn}}!}{(s!)^E}$ possible balanced partitions. Furthermore, assignment and routing are tightly coupled, i.e., changing which neurons belong to an expert changes what inputs should be routed to it, and vice versa. This interdependence precludes simple two-stage approaches that fix assignments before training routers.

We propose to \emph{jointly learn} the neuron-to-expert assignment and the token-to-expert routing by formulating assignment as an optimal transport problem. The key insight is that neuron assignment can be viewed as \emph{transporting mass} from neurons to experts: each neuron carries unit mass to be delivered to exactly one expert, while each expert must receive exactly $s$ units. The cost of each assignment is determined by how well the resulting MoE reconstructs the dense FFN output. This perspective maps directly to optimal transport (OT), which finds minimum-cost mass redistribution under marginal constraints.

This formulation requires three components:
\begin{enumerate}[noitemsep]
    \item A \textit{differentiable} relaxation of the discrete assignment that permits gradient-based optimization.
    \item \textit{Hard capacity constraints} ensuring each expert receives exactly $s$ neurons and each neuron is assigned to exactly one expert.
    \item An \textit{output-aware objective} that directly measures deviation from the dense FFN output, not intermediate representations.
\end{enumerate}

\subsection{Neuron-to-Expert Assignment via Optimal Transport}

We first formalize the assignment problem using the language of optimal transport.

\begin{definition}[Optimal Transport Problem]
\label{def:ot}
Given source distribution $\mathbf{r} \in \mathbb{R}^m_+$ and target distribution $\mathbf{c} \in \mathbb{R}^n_+$ with $\sum_i r_i = \sum_j c_j$, and a cost matrix $\mathbf{C} \in \mathbb{R}^{m \times n}$, the optimal transport problem seeks a transport plan $\mathbf{M}^*$ that minimizes total transportation cost:
\begin{equation}
\mathbf{M}^* = \underset{\mathbf{M} \in \mathcal{U}(\mathbf{r}, \mathbf{c})}{\argmin} \langle \mathbf{C}, \mathbf{M} \rangle
\label{eq:ot_general}
\end{equation}
where $\mathcal{U}(\mathbf{r}, \mathbf{c}) = \{\mathbf{M} \geq 0 : \mathbf{M}\mathbf{1}_n = \mathbf{r}, \mathbf{M}^\top\mathbf{1}_m = \mathbf{c}\}$ is the set of valid transport plans (the \emph{transportation polytope}), and $\langle \cdot, \cdot \rangle$ denotes the Frobenius inner product.
\end{definition}

For neuron assignment, we set $m = d_{\text{ffn}}$ (neurons) and $n = E$ (experts), with marginals $\mathbf{r} = \mathbf{1}_{d_{\text{ffn}}}$ (each neuron assigned exactly once) and $\mathbf{c} = s \cdot \mathbf{1}_E$ (each expert receives exactly $s$ neurons). Rather than specifying a fixed cost matrix, we introduce a \emph{learnable affinity matrix} $\mathbf{A} \in \mathbb{R}^{d_{\text{ffn}} \times E}$ where $A_{i,e}$ represents the affinity of assigning neuron $i$ to expert $e$. Setting the cost as $\mathbf{C} = -\mathbf{A}$, our objective becomes:
\begin{equation}
\mathbf{M}^* = \underset{\mathbf{M} \in \mathcal{U}(\mathbf{r}, \mathbf{c})}{\text{argmin}} \langle -\mathbf{A}, \mathbf{M} \rangle = \underset{\mathbf{M} \in \mathcal{U}(\mathbf{r}, \mathbf{c})}{\text{argmax}} \langle \mathbf{A}, \mathbf{M} \rangle
\label{eq:ot_problem}
\end{equation}
This seeks the assignment that maximizes total affinity while satisfying the balance constraints.

\textbf{Limitation of Standard OT. }While Eq.~\ref{eq:ot_problem} captures our desired solution, $\mathbf{M}^*$ lies at a vertex of the transportation polytope which is a $\{0,1\}$-matrix where each row contains exactly one entry equal to 1 and each column sums to $s$. This discrete structure presents two challenges: (1) the $\argmin$ over a polytope is non-differentiable, blocking gradient flow to the affinity matrix $\mathbf{A}$, and (2) solving the linear program exactly at each training step is computationally prohibitive.

\textbf{Entropic Regularization. }To obtain a differentiable solution, we add entropic regularization to the OT objective~\cite{cuturi2013sinkhorndistanceslightspeedcomputation}:
\begin{equation}
\mathbf{M}^*_\tau = \underset{\mathbf{M} \in \mathcal{U}(\mathbf{r}, \mathbf{c})}{\argmin} \langle -\mathbf{A}, \mathbf{M} \rangle - \tau H(\mathbf{M})
\label{eq:entropic_ot}
\end{equation}
where $H(\mathbf{M}) = -\sum_{i,e} M_{i,e}(\log M_{i,e} - 1)$ is the entropy of the transport plan and $\tau > 0$ is a temperature parameter. The entropy term $-\tau H(\mathbf{M})$ strictly convexifies the objective, yielding a unique solution in the \emph{interior} of the polytope rather than at a vertex. As $\tau \to 0$, the solution approaches the unregularized optimum; as $\tau \to \infty$, it approaches the uniform plan.

The key advantage of entropic regularization is that the solution admits a closed-form factorization:
\begin{equation}
M^*_{i,e} = u_i \cdot \exp(A_{i,e}/\tau) \cdot v_e
\label{eq:sinkhorn_form}
\end{equation}
where scaling vectors $\mathbf{u} \in \mathbb{R}^{d_{\text{ffn}}}_+$ and $\mathbf{v} \in \mathbb{R}^E_+$ can be found via the Sinkhorn-Knopp algorithm~\citep{sinkhorn1967concerning, knight2008sinkhorn}, which performs alternating row and column normalizations that converge linearly to the unique solution satisfying the marginal constraints. We denote the resulting soft assignment as $\mathbf{M}_{\text{soft}} \in [0,1]^{d_{\text{ffn}} \times E}$, computed via log-domain Sinkhorn iterations for numerical stability (see Appendix~\ref{app:log_domain_sinkhorn}).

\subsection{Token-to-Expert Routing}
In addition to assigning neurons to experts, we must learn which experts to activate for each input token. We parameterize the router as a linear projection followed by top-$k$ selection.

For input tokens $\mathbf{X} \in \mathbb{R}^{n \times d}$, the router computes:
\begin{align}
\mathbf{L} &= \mathbf{X}\mathbf{W}_{\text{r}}^\top \in \mathbb{R}^{n \times E} \label{eq:router_logits}\\
\mathbf{P} &= \softmax(\mathbf{L}) \in \mathbb{R}^{n \times E} \label{eq:router_probs}\\
\mathcal{I}_i &= \topk(\mathbf{P}_i, k) \quad \forall i \in \{1, \ldots, n\} \label{eq:router_topk}
\end{align}
where $\mathbf{W}_{\text{r}} \in \mathbb{R}^{E \times d}$ are learnable router weights and $\mathcal{I}_i$ contains the indices of the $k$ experts selected for token $i$.

\subsection{Differentiable Assignment and Routing}
With the affinity matrix $\mathbf{A}$ governing neuron-to-expert assignment and router weights $\mathbf{W}_r$ governing token-to-expert routing, we now describe how to jointly optimize both. Two challenges must be addressed: (1) converting the soft assignment $\mathbf{M}_{\text{soft}}$ to discrete expert clusters, and (2) enabling gradient flow through both the discrete assignment and the top-$k$ routing selection.

\textbf{Hard Assignment via Greedy Rounding.} The soft assignment $\mathbf{M}_{\text{soft}}$ provides fractional neuron-to-expert allocations, but deployment requires discrete assignments. We convert $\mathbf{M}_{\text{soft}}$ to a binary matrix $\mathbf{M} \in \{0,1\}^{d_{\text{ffn}} \times E}$ via greedy selection: sort all entries of $\mathbf{M}_{\text{soft}}$ in descending order, then iteratively assign neuron $i$ to expert $e$ if neuron $i$ is unassigned and expert $e$ has capacity remaining. This yields disjoint expert clusters $\mathcal{C}_1, \ldots, \mathcal{C}_E$ with $|\mathcal{C}_e| = s$. A potential concern is mismatch between soft and hard assignments, e.g., when a neuron's preferred expert has reached capacity. However, Sinkhorn already accounts for capacity constraints globally, redistributing probability mass when experts are over-demanded.

\textbf{Gradient Estimation via STE.} Both the greedy rounding for assignment and the top-$k$ selection for routing are non-differentiable. We employ straight-through estimators (STE)~\citep{bengio2013estimating} that use hard decisions in the forward pass while routing gradients through the soft counterparts in the backward pass:
\begin{align}
\mathbf{M}_{\text{STE}} &= \mathbf{M} + (\mathbf{M}_{\text{soft}} - \sg(\mathbf{M}_{\text{soft}})) \label{eq:assignment_ste}\\
\mathbf{R}_{\text{STE}} &= \mathbf{R} + (\mathbf{P} - \sg(\mathbf{P})) \label{eq:router_ste}
\end{align}
where $\sg(\cdot)$ denotes the stop-gradient operator, $\mathbf{M} \in \{0,1\}^{d_{\text{ffn}} \times E}$ is the hard neuron assignment from greedy rounding, and $\mathbf{R} \in \{0,1\}^{n \times E}$ is the binary routing mask with $R_{i,e} = 1$ iff expert $e$ is selected for token $i$. This allows end-to-end optimization: gradients from the reconstruction loss flow through $\mathbf{M}_{\text{soft}}$ to update $\mathbf{A}$, and through $\mathbf{P}$ to update $\mathbf{W}_r$.

\subsection{Alignment Phase}

During training, we simulate sparse MoE computation by masking the intermediate activations: $\hat{\mathbf{Y}} = (\mathbf{H} \odot (\mathbf{R}\mathbf{M}^{\top})) \mathbf{W}_{\text{down}}$, where only the $k \cdot s$ neurons belonging to the selected experts contribute to each token's output (see Appendix~\ref{app:sparse_moe_computation}). We jointly optimize the assignment logits $\mathbf{A}$ and router weights $\mathbf{W}_{\text{r}}$ across the network using a combination of losses aimed at preserving the residual stream: KL divergence between the pretrained dense teacher and MoE student output distributions, cross-entropy loss on the language modeling objective, and auxiliary losses for MoE training stability. Specifically, we employ router z-loss~\citep{zoph2022st} to penalize large router logits and prevent instability, and load balancing loss~\citep{shazeer2017sparsely} to encourage uniform expert utilization and prevent expert collapse. Full details are provided in Appendix~\ref{app:training_objective}.

Once training converges, we extract the final assignment $\mathbf{M}$ and convert the model into a standard MoE architecture with $E$ distinct expert FFNs, enabling efficient sparse inference. The same balanced-transport formulation extends directly to multi-head attention by grouping heads into experts; we defer the full formulation and results to Appendix~\ref{sec:attention_moe}.
\begin{table}[t]
\centering
\caption{Comparison of Perplexity on WikiText-2 and HellaSwag at 50\% parametric budget. DOT-MoE outperforms existing structured and semi-structured methods on LLaMA-2 7B.}
\label{tab:llama7b_comparison}
\setlength{\tabcolsep}{10pt}
\begin{tabular}{l cc}
\toprule
\multirow{2}{*}{\textbf{Method}} & \multicolumn{2}{c}{\textbf{LLaMA-2 7B}} \\
\cmidrule(lr){2-3}
 & \scriptsize{WikiText [PPL $\downarrow$]} & \scriptsize{HellaS. [acc-n $\uparrow$]} \\
\midrule
\rowcolor{gray!10} \multicolumn{3}{c}{\textit{Structured Pruning (50\%)}} \\
LLM-Pruner   & 31.05  & -- \\
LLM Surgeon  & 15.38  & 40.3 \\
ShortGPT     & 268.11 & 43.7 \\
SLEB         & 103.38 & -- \\
K-OBD        & 46.64 & 36.8 \\
SliceGPT     & 24.82  & 33.0 \\
ModeGPT      & 11.88  & -- \\
DISP-LLM     & 9.84   & 46.3 \\
\midrule
\rowcolor{gray!10} \multicolumn{3}{c}{\textit{Semi-Structured Pruning (2:4)}} \\
SparseGPT    & 10.17  & 43.3 \\
Wanda        & 11.02  & 40.9 \\
Pruner-Zero  & 10.52  & \textbf{54.7} \\
\midrule
\rowcolor{blue!5} \textbf{DOT-MoE} & \textbf{7.99} & 53.9 \\
\bottomrule
\end{tabular}
\end{table}

\section{Experiments}
\begin{table*}[h]
\caption{Fine-tuning performance comparison on common-sense reasoning benchmarks. \#FT Tokens denotes the fine-tuning data budget after conversion. Dense rows report the original pretraining budget (marked with $^*$) for reference; no fine-tuning is applied to dense models. DOT-MoE models recover performance with fewer tokens and bridge the gap to dense counterparts.}
\label{tab:fine_tuned_results}
\centering
\setlength{\tabcolsep}{4pt}
\begin{tabular}{l c c cccccc c}
\toprule
\multirow{2}{*}{\textbf{Method}} & \textbf{Active} & \multirow{2}{*}{\textbf{\#FT Tokens}} & \textbf{BoolQ} & \textbf{SciQ} & \textbf{PIQA} & \textbf{WinoG.} & \textbf{ARC-C} & \textbf{HellaS.} & \multirow{2}{*}{\textbf{Avg.}} \\
 & \textbf{Params} & & \scriptsize{[32, acc]} & \scriptsize{[0, acc]} & \scriptsize{[0, acc]} & \scriptsize{[5, acc]} & \scriptsize{[25, acc-n]} & \scriptsize{[10, acc-n]} & \\
\midrule
\rowcolor{gray!10} \multicolumn{10}{c}{\textit{LLaMA-2 7B}} \\
Dense & 6.74B & 2T$^*$ & 82.0 & 94.0 & 78.1 & 74.3 & 52.5	& 78.9 & 76.6 \\
LLaMA-MoE (E8A2)$^\dagger$ & 3.49B & 1.2B & 37.8 & 20.0 & 49.7 & 50.1 & 25.8 & 26.2 & 34.9 \\
LLaMA-MoE-v2 (E8A2)$^\dagger$ & 3.49B & 1.2B & 51.3 & 67.0 & 56.6 & 52.9 & 25.7 & 35.1 & 48.1 \\
CMoE (E8S1A1)$^\dagger$ & 3.49B & 1.2B & 55.0 & 77.5 & 57.1 & 54.1 & 27.6 & 38.8 & 51.7 \\
\rowcolor{blue!5} \textbf{DOT-MoE } & 3.49B & 1.2B & \textbf{72.5} & \textbf{94.3} & \textbf{69.3} & \textbf{62.5} & \textbf{40.9} & \textbf{60.2} & \textbf{66.6} \\
\midrule
\rowcolor{gray!10} \multicolumn{10}{c}{\textit{LLaMA-3 8B}} \\
Dense & 8.03B & 15T$^*$ & 83.2 & 96.2 & 79.6 & 77.3 & 58.3 & 82.1 & 79.4 \\
CMoE (E8A1S1) & 3.80B & 1.2B & 71.1 & 94.4 & 69.5 & 59.5 & 38.2 & 55.3 & 64.7 \\
\rowcolor{blue!5} \textbf{DOT-MoE } & 3.80B & 1.2B & 75.0 & 94.2 & 70.2 & 63.8 & 42.4 & 61.1 & 67.8 \\
LLaMA-MoE-v2 (E8A2)$^\ddagger$ & 3.80B & 7B & 74.6 & 94.5 & 69.3 & 60.5 & {42.8} & 59.0 & 66.8 \\
LLaMA-MoE-v2 (E8A1S1)$^\ddagger$ & 3.80B & 7B & \textbf{76.9} & 92.8 & 67.9 & 58.4 & 40.2 & 53.7 & 65.0 \\
\rowcolor{blue!5} \textbf{DOT-MoE } & 3.80B & 7B & 75.4 & \textbf{96.2} & \textbf{73.3} & \textbf{66.1} & \textbf{49.1} & \textbf{66.0} & \textbf{71.0} \\
\midrule
\rowcolor{gray!10} \multicolumn{10}{c}{\textit{Qwen2.5 7B}} \\
Dense & 7.62B & 18T$^*$ & 88.0 & 96.6 & 78.6 & 76.2 & 64.1 & 80.3 & 80.6 \\
CMoE (E8A1S1) & 3.34B & 1.2B & 69.4 & 93.8 & 67.6 & 59.0 & 37.8 & 57.2 & 64.1 \\
\rowcolor{blue!5} \textbf{DOT-MoE } & 3.34B & 1.2B & 75.2 & \textbf{95.0} & 70.4 & 61.3 & 42.8 & 58.4 & 67.2 \\
\rowcolor{blue!5} \textbf{DOT-MoE} & 4.76B & 1.2B & \textbf{81.3} & 94.6 & \textbf{73.1} & \textbf{68.1} & \textbf{49.8} & \textbf{67.3} & \textbf{73.4} \\
\bottomrule
\multicolumn{10}{l}{\footnotesize $^\dagger$Results from \cite{cmoe}. $^\ddagger$Results from \cite{llamamoev2}.}
\end{tabular}%
\end{table*}

\subsection{Experimental Setup}

\textbf{Models and Evaluation.} We evaluate on three publicly available dense checkpoints: LLaMA-2-7B~\cite{touvron2023llama}, LLaMA-3-8B~\cite{grattafiori2024llama}, and Qwen2.5-7B~\cite{qwen2.5}. All methods are evaluated using lm-evaluation-harness~\cite{eval-harness} with benchmark’s default prompts and standard few-shot settings: ARC-Challenge (25-shot)~\cite{clark2018think}, Winogrande (5-shot)~\cite{sakaguchi2021winogrande}, HellaSwag (10-shot)~\cite{zellers2019hellaswag}, PIQA (0-shot)~\cite{bisk2020piqa}, SciQ (0-shot)~\cite{welbl2017crowdsourcingsciq}, and BoolQ (32-shot)~\cite{clark2019boolq}.

\textbf{Implementation.} DOT-MoE is implemented using PyTorch~\cite{paszke2019pytorch} and Hugging Face Transformers~\cite{wolf2020transformers}. We \textbf{freeze} the dense model weights during the alignment phase and only train the assignment logits $\mathbf{A}$ and router weights $\mathbf{W}_r$. Unless stated otherwise, each expert contains $s=128$ intermediate neurons, yielding $E=148, 112, \text{and } 86$ experts per layer for Qwen2.5-7B, LLaMA-3-8B, and LLaMA-2-7B, respectively. We use top-$k$ routing with a fixed fraction of active experts per token. For our main experiments, we target 25\% FFN sparsity which translates to $k=37, 28 \text{and }22$ active experts for Qwen2.5-7B, LLaMA-3-8B, and LLaMA-2-7B respectively. We use AdamW with cosine learning rate decay and linear warmup. All experiments are conducted on 8xH100 GPUs. Full hyperparameters are provided in Appendix~\ref{app:implementation_details}.

\textbf{Training Data.} We use Dolmino-mix~\cite{olmo20252olmo2furious} for the alignment phase as well as for continuous fine-tuning. Alignment is done for 3500 steps which takes $<3$ hours on 8xH100 GPUs for the LLaMA3-8B; a detailed profiling of Sinkhorn and STE overhead is provided in Appendix~\ref{app:training_overhead}. We exclusively train the assignment logits $\mathbf{A}$ and router weights $\mathbf{W}_{\text{r}}$ in the alignment phase, completely freezing the dense model weights. In order to be consistent with existing works, we further use 1.2B tokens for continuous fine-tuning of the aligned model. We sample the same data from the larger Dolmino-mix for training DOT-MoE as well as the baselines.

\textbf{Baselines.} We compare DOT-MoE against a broad set of structured pruning, semi-structured pruning and dense-to-MoE conversion methods. Structured pruning baselines include LLM-Pruner \cite{ma2023llmpruner}, LLM Surgeon \cite{van2023llmsurgeon}, ShortGPT \cite{men2025shortgpt}, SLEB \cite{song2024sleb}, K-OBD,  SliceGPT \cite{ashkboos2024slicegpt}, ModeGPT \cite{lin2024modegpt} and DISP-LLM \cite{gao2024disp}. Semi-structured pruning baselines include Wanda \cite{sunsimple}, SparseGPT \cite{frantar2023sparsegpt} and Pruner-Zero \cite{dong2024pruner}.  Dense-to-MoE baselines include CMoE \cite{cmoe}, LLaMA-MoE \cite{llamamoe} and LLaMA-MoE-v2 \cite{llamamoev2}. All baselines are evaluated under comparable sparsity. We justify baseline selection and provide additional comparisons in Appendix~\ref{app:additional_baselines}.

\begin{figure*}[t!]
    \centering
    \begin{subfigure}[t]{0.32\textwidth}
        \centering
        \includegraphics[width=\linewidth]{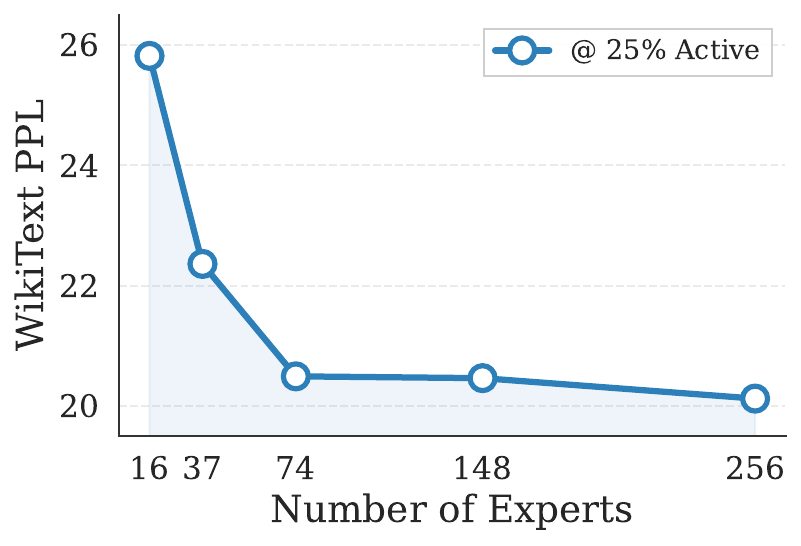}
        \caption{Expert Granularity}
        \label{fig:expert_granularity}
    \end{subfigure}
    \hfill
    \begin{subfigure}[t]{0.32\textwidth}
        \centering
        \includegraphics[width=\linewidth]{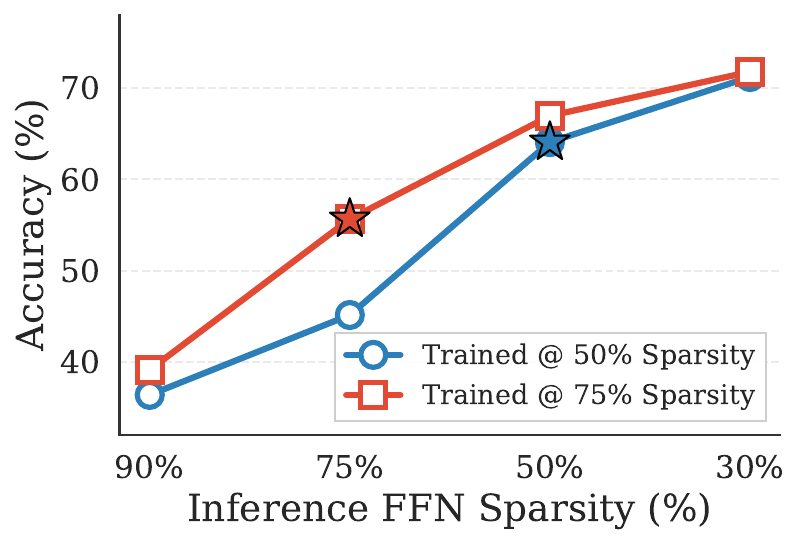}
        \caption{Effect of training sparsity}
        \label{fig:experts_per_token}
    \end{subfigure}
    \hfill
    \begin{subfigure}[t]{0.32\textwidth}
      \centering
      \includegraphics[width=\linewidth]{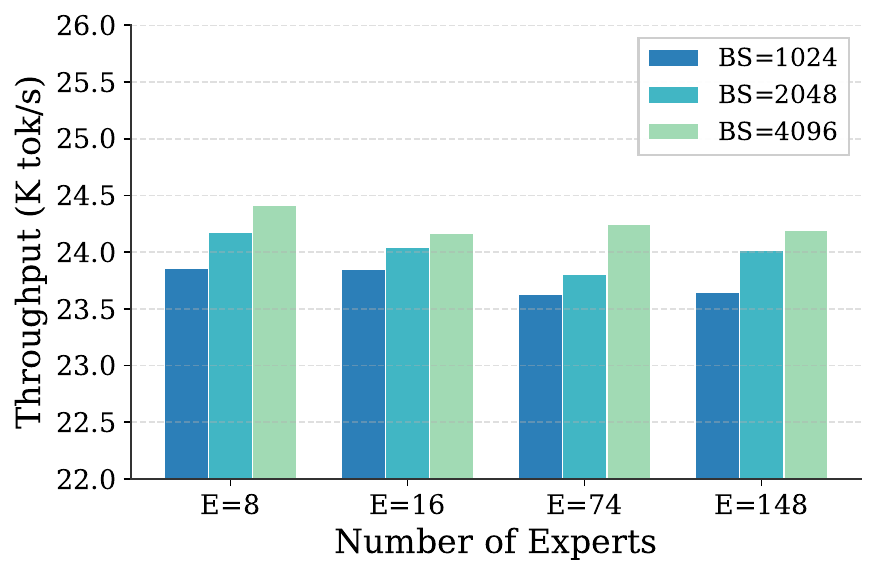}
      \caption{Inference Throughput}
      \label{fig:throughput}
    \end{subfigure}
    \caption{Ablation results for \name. (a) Increasing expert granularity improves performance until saturation. (b) Training with higher FFN sparsity yields robust expert representations that generalize better to extreme sparsity regimes at inference time. (c) Inference throughput remains stable across expert granularities when active parameters are held constant.}
    \label{fig:ablation_overview}
\end{figure*}

\subsection{Main Results}
\begin{table*}[t]
\caption{Zero-shot performance comparison on standard common-sense reasoning benchmarks. DOT-MoE consistently outperforms existing structured pruning and dense-to-MoE conversion methods across multiple model families.}
\label{tab:zero_shot_results}
\centering
\setlength{\tabcolsep}{4pt}
\begin{tabular}{l c cccccc c}
\toprule
\multirow{2}{*}{\textbf{Method}} & \textbf{Active} & \textbf{BoolQ} & \textbf{SciQ} & \textbf{PIQA} & \textbf{WinoG.} & \textbf{ARC-C} & \textbf{HellaSwag} & \multirow{2}{*}{\textbf{Avg.}} \\
 & \textbf{Params} & \scriptsize{[32, acc]} & \scriptsize{[0, acc]} & \scriptsize{[0, acc]} & \scriptsize{[5, acc]} & \scriptsize{[25, acc-n]} & \scriptsize{[10, acc-n]} & \\
\midrule
\rowcolor{gray!10} \multicolumn{9}{c}{\textit{LLaMA-2 7B}} \\
SliceGPT & 3.73B & 38.3 & 73.4 & 55.2 & 52.8 & 24.5 & 33.0 & 46.2 \\
DISP-LLM & 3.57B & 60.0 & 83.5 & 65.2 & 57.0 & 29.4 & 49.5 & 57.4 \\
CMoE (E8A1S1) & 3.49B & 46.1 & 65.3 & 52.8 & 48.7 & 23.8 & 30.1 & 44.5 \\
\rowcolor{blue!5} \textbf{DOT-MoE } & 3.49B & \textbf{66.4} & \textbf{89.6} & \textbf{67.2} & \textbf{57.9} & \textbf{34.0} & \textbf{53.9} & \textbf{61.5} \\
\midrule
\rowcolor{gray!10} \multicolumn{9}{c}{\textit{LLaMA-3 8B}} \\
SliceGPT & 3.90B & 37.8 & 62.4 & 53.3 & 49.6 & 22.1 & 27.6 & 42.1 \\
DISP-LLM & 3.84B & 51.0 & 84.5 & 60.6 & 53.9 & 26.3 & 40.1 & 52.7 \\
CMoE (E8A1S1) & 3.80B & 41.0 & 51.1 & 53.9 & 51.0 & 25.3 & 28.6 & 41.8 \\
\rowcolor{blue!5} \textbf{DOT-MoE } & 3.80B & \textbf{62.3} & \textbf{90.9} & \textbf{65.0} & \textbf{56.0} & \textbf{34.1} & \textbf{50.1} & \textbf{59.8} \\
\midrule
\rowcolor{gray!10} \multicolumn{9}{c}{\textit{Qwen2.5 7B}} \\
ShortGPT & 4.82B & 42.2 & 55.5 & 59.4 & 51.7 & 27.5 & 35.9 & 45.4 \\
SliceGPT & 4.75B & 38.0 & 80.6 & 56.5 & 53.5 & 22.8 & 33.5 & 47.5 \\
DISP-LLM & 4.91B & 65.7 & 94.1 & 70.2 & 62.8 & 45.9 & 61.3 & 66.7 \\
CMoE (E8A2S2) & 4.76B & 60.3 & 87.0 & 62.8 & 51.5 & 28.4 & 42.7 & 55.5 \\
\rowcolor{blue!5} \textbf{DOT-MoE} & 4.76B & \textbf{80.1} & \textbf{95.0} & \textbf{74.2} & \textbf{66.2} & \textbf{50.3} & \textbf{67.9} & \textbf{72.3} \\
\midrule
DISP-LLM  &  3.67B  & 46.8&	84.8 &	63.5 &	57.2	&35.1&	46.4	& 55.6 \\
CMoE (E8A1S1) & 3.34B & 43.1	&41.8	&54.2	&50.4&	25.0	&27.1	&40.3 \\
\rowcolor{blue!5} \textbf{DOT-MoE} & \textbf{3.34B} & \textbf{64.3}	&\textbf{93.9}&	\textbf{68.5}&	\textbf{58.7}	&\textbf{39.6}&	\textbf{55.0}	&\textbf{63.3} \\
\bottomrule
\end{tabular}%
\end{table*}

\paragraph{Comparison with Pruning Methods.} 
Table~\ref{tab:llama7b_comparison} compares DOT-MoE against structured and semi-structured pruning methods on LLaMA-2 7B at 50\% parametric budget. DOT-MoE achieves the lowest perplexity (7.99) among all existing methods, outperforming the state-of-the-art DISP-LLM (9.84) by a substantial margin. DOT-MoE is also competitive with semi-structured pruning methods which have a greater degree of freedom to achieve any target sparsity. 

This advantage extends to downstream tasks. As shown in Table~\ref{tab:zero_shot_results}, DOT-MoE demonstrates superior knowledge retention compared to pruning baselines across all three model families. On Qwen2.5-7B, DOT-MoE outperforms DISP-LLM (72.3\% vs 66.7\% average accuracy), confirming that dense-to-MoE conversion is more effective than pruning, as it preserves total model capacity while activating only a subset of parameters per token.

\paragraph{Comparison with Dense-to-MoE Methods.} 
We compare DOT-MoE with existing dense-to-MoE conversion methods in Table~\ref{tab:zero_shot_results}. A key distinction is that methods like LLaMA-MoE and CMoE first permanently assign neurons to experts, then train a randomly initialized router on this fixed partition. This requires extensive fine-tuning to recover performance. In contrast, DOT-MoE jointly learns neuron assignment and routing during the alignment phase, allowing them to co-adapt. This enables a stronger \textit{zero-shot} transfer without training the model weights at all. As shown in Table~\ref{tab:zero_shot_results}, DOT-MoE achieves 61.5\% average accuracy on LLaMA-2 7B, substantially outperforming CMoE (44.5\%). A similar trend can be seen for other models. These results demonstrate that our output-aware expert construction optimizes the trade-off between sparsity and reconstruction fidelity more effectively than activation-based clustering. 

\paragraph{Impact of Fine-tuning.} 
While DOT-MoE is effective out-of-the-box, we investigate the impact of continuous fine-tuning. As shown in Table~\ref{tab:fine_tuned_results}, fine-tuning DOT-MoE on 1.2B tokens boosts LLaMA-2 7B accuracy from 61.5\% to 66.6\%, widening the gap against CMoE (51.7\%) and LLaMA-MoE-v2 (48.1\%). On LLaMA-3 8B, DOT-MoE achieves 67.8\% with 1.2B tokens and improves to 71.0\% when scaled to 7B tokens, outperforming LLaMA-MoE-v2 (66.8\%) trained on the exact similar data split. This scaling behavior confirms that DOT-MoE provides a superior initialization for sparse models that continues to benefit from additional training data. It is worth noting that DOT-MoE substantially closes the gap between Dense and Sparse model with as few as 1.2B tokens of training. On Qwen2.5-7B DOT-MoE achieves an average accuracy of 73.4\% whereas the dense pre-trained model has an average accuracy of 80.6\%. DOT-MoE also scales to larger models: on Qwen2.5-32B it improves over CMoE by +34.3 points, and maintains consistent gains across context lengths up to 32K tokens (Appendix~\ref{app:scalability}). It would be interesting to study the scaling behavior of dense-to-MoE models in the $\mathtt{\sim}$100B token regime but we leave that to future work due to compute constraints.

\begin{figure*}[t!]
    \centering
    \begin{subfigure}[t]{0.32\textwidth}
        \centering
        \includegraphics[width=\linewidth]{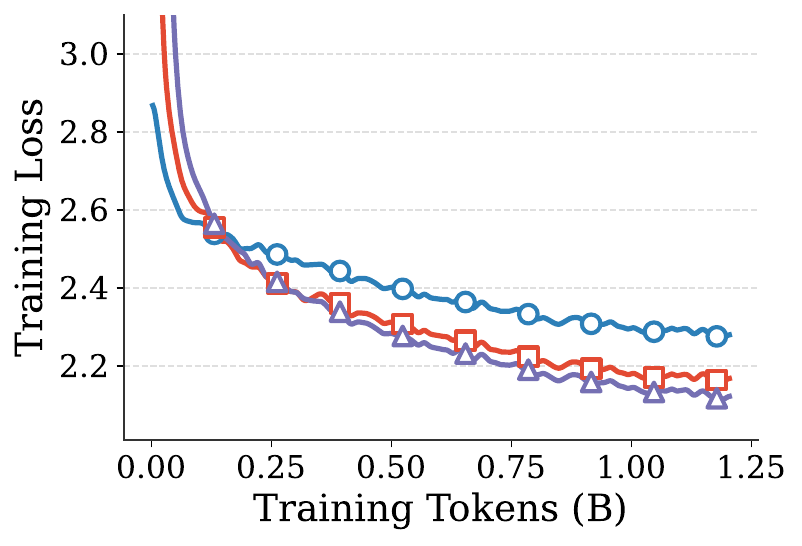}
        \caption{Training Loss}
        \label{fig:training_curve}
    \end{subfigure}
    \hfill
    \begin{subfigure}[t]{0.32\textwidth}
        \centering
        \includegraphics[width=\linewidth]{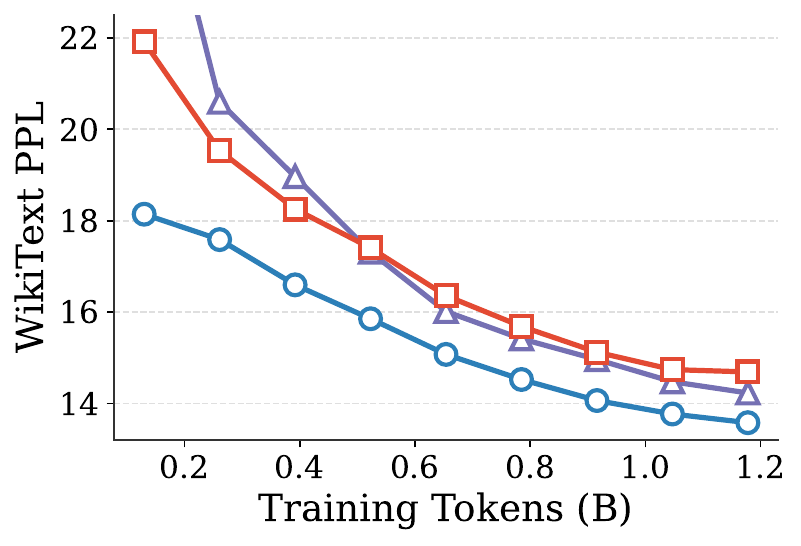}
        \caption{WikiText PPL}
        \label{fig:validation_curve}
    \end{subfigure}
    \hfill
    \begin{subfigure}[t]{0.32\textwidth}
      \centering
      \includegraphics[width=\linewidth]{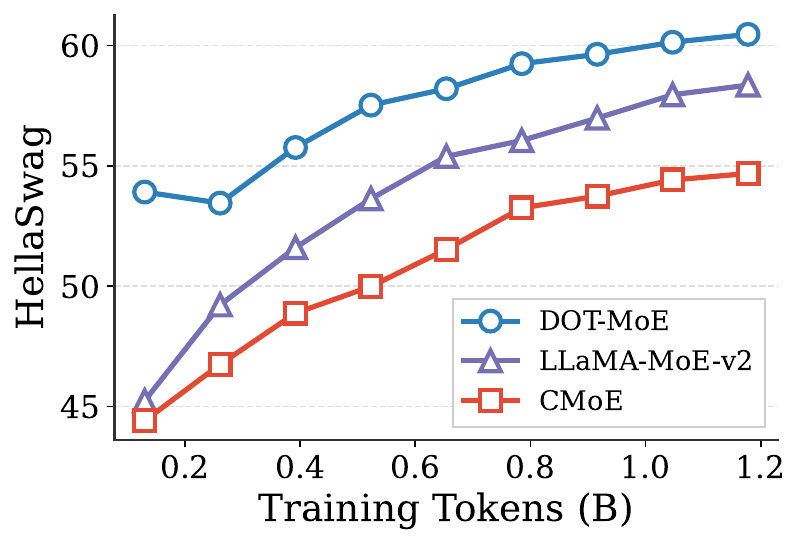}
      \caption{HellaSwag Acc Norm}
      \label{fig:hellaswag_curve}
    \end{subfigure}
    \caption{Effect of initialization on training dynamics. DOT-MoE starts with substantially lower training loss and WikiText perplexity, maintaining this advantage throughout fine-tuning. This translates to consistently higher downstream accuracy on HellaSwag.}
    \label{fig:effect_of_init}
\end{figure*}

\subsection{Ablation Studies}

\subsubsection{Expert Granularity}
In Figure~\ref{fig:expert_granularity}, we investigate the effect of expert granularity on performance by varying the total number of experts ($E \in \{16, 37, 74, 148, 256\}$) in DOT-MoE on Qwen2.5 7B. One might argue that DOT-MoE exhibits better performance due to a higher number of experts. However prior Dense-to-MoE methods~\cite{cmoe, llamamoev2} observed that increasing expert count from 8 to 16 leads to minimal improvement or degraded performance due to increased routing complexity, whereas DOT-MoE maintains stable performance at much higher expert counts. To test this, we trained a CMoE model with Qwen2.5-7B  backbone. We set the total number of experts to 37 and active experts to 9; and observed a $>$5K WikiText perplexity for CMoE. In case of DOT-MoE, increasing the number of experts initially improves performance, but gains saturate as granularity becomes excessively fine, consistent with observations in OLMoE~\cite{muennighoff2025olmoeopenmixtureofexpertslanguage}. To control for the granularity confound, we also run DOT-MoE at CMoE's own setting ($E{=}8$, top-$k{=}2$) and still observe consistent gains across three model families (Appendix~\ref{app:same_granularity}). 

\begin{tcolorbox}[
    colback=findingbg,      
    colframe=findingborder, 
    arc=3mm,                
    boxrule=0.8pt,          
    left=4pt, right=4pt, top=4pt, bottom=4pt, 
    boxsep=0pt
]
    \textbf{Observation 1:} Routing benefits from fine-grained experts up to a point beyond which additional experts provide limited returns.
\end{tcolorbox}

\subsubsection{Effect of Expert Granularity on Inference Speed}
A natural concern with fine-grained experts is inference overhead: does increasing the number of experts slow down generation? To investigate, we benchmark inference throughput using vLLM's fused MoE kernels~\cite{kwon2023efficient} across four expert configurations ($E \in \{8, 16, 74, 148\}$) while holding active parameters constant at 25\% of the FFN.

Figure~\ref{fig:throughput} shows throughput (tokens/sec) across batch sizes. Crucially, throughput remains stable as expert count increases. This is because vLLM's fused MoE implementation batches all expert computations into a small number of large GEMMs rather than executing each expert separately. The fused weights are stored as $\mathbf{W}_{\text{fused}} = [\mathbf{W}_1, \ldots, \mathbf{W}_E]$ along the expert dimension, and token reordering enables a single large matrix multiplication regardless of expert count. Since the total fused intermediate dimension ($E \times s$) and active neurons per token ($k \times s$) remain constant, the GEMM sizes and thus throughput are largely unaffected by expert granularity.

\begin{tcolorbox}[
    colback=findingbg,
    colframe=findingborder,
    arc=3mm,
    boxrule=0.8pt,
    left=4pt, right=4pt, top=4pt, bottom=4pt,
    boxsep=0pt
]
    \textbf{Observation 2:} Fine-grained experts incur no throughput penalty with fused MoE kernels when active parameters are held constant.
\end{tcolorbox}

\subsubsection{Effect of Training Sparsity.}
A key advantage of our method is the ability to dynamically adjust the number of active experts at inference time, enabling flexible compute-accuracy trade-offs without alignment or fine-tuning. We investigate how the sparsity level used during training affects the model's generalization across different inference-time sparsity configurations. We train two Qwen2.5-7B models with different FFN sparsity levels (50\% and 75\%) and evaluate both across a range of inference sparsities (30\%, 50\%, 75\%, 90\%). Figure~\ref{fig:experts_per_token} reports the average accuracy across the same six benchmarks reported in Table~\ref{tab:zero_shot_results}as a function of FFN sparsity at inference time. Individual benchmark results are present in the Appendix Table~\ref{tab:sparsity_detailed}. We observe an interesting behavior - models aligned with a higher sparsity level yields more robust expert representations across inference sparsities. For example, the model trained at 75\% sparsity consistently outperforms the 50\% trained model across varying inference sparsities. This behavior can be explained by our reconstruction objective. When trained with fewer active experts, the model learns to encode information more efficiently within each expert, resulting in more compact and discriminative representations.

\begin{tcolorbox}[
    colback=findingbg,      
    colframe=findingborder, 
    arc=3mm,                
    boxrule=0.8pt,          
    left=4pt, right=4pt, top=4pt, bottom=4pt, 
    boxsep=0pt
]
    \textbf{Observation 3:} Training at higher sparsity yields experts that generalize better across varying inference sparsities.
\end{tcolorbox}

\subsubsection{Effect of Initialization on Training Dynamics}
We investigate how different expert construction strategies affect fine-tuning dynamics by comparing DOT-MoE, CMoE, and LLaMA-MoE-v2 on LLaMA-3 8B. All methods are trained on the same data split at 25\% FFN sparsity. Figure~\ref{fig:effect_of_init} shows training loss, WikiText perplexity, and zero-shot HellaSwag normalized accuracy as a function of training tokens.

DOT-MoE exhibits a clear advantage at initialization, starting with substantially lower training loss compared to CMoE and LLaMA-MoE-v2. While all methods reduce training loss over time, CMoE and LLaMA-MoE-v2 exhibit signs of overfitting: despite achieving lower training loss, they consistently show higher WikiText perplexity and lower HellaSwag accuracy compared to DOT-MoE. In contrast, DOT-MoE continues to improve on both validation perplexity and downstream task performance throughout training, demonstrating that its expert construction leads to more distinct and generalizable representations.

These results show that output-aware expert construction provides not only a better starting point but also a more robust learning potential. The quality of the initial expert assignment directly impacts generalization, validating our approach of jointly optimizing neuron assignment and routing during the alignment phase.

\begin{tcolorbox}[
    colback=findingbg,      
    colframe=findingborder, 
    arc=3mm,                
    boxrule=0.8pt,          
    left=4pt, right=4pt, top=4pt, bottom=4pt, 
    boxsep=0pt
]
    \textbf{Observation 4:} Output-aware initialization achieves superior training generalization whereas heuristic methods exhibit overfitting.
\end{tcolorbox}

\section{Future Work and Conclusion}
In this work, we introduced DOT-MoE, a novel framework that formulates the conversion of dense FFNs into sparse MoEs as a Differentiable Optimal Transport problem. By jointly learning the neuron-to-expert assignment and the routing policy via Straight-Through Estimators, we achieve a superior trade-off between sparsity and performance compared to heuristic clustering or structured pruning methods. We further show that the same balanced-transport formulation generalizes to multi-head attention, yielding substantial gains when applied to attention-head assignment. Several promising directions remain for future research. First, while we currently initialize the affinity matrix $\mathbf{A}$ randomly, exploring data-driven initializations such as leveraging weight correlations or pre-computed activation statistics could accelerate Sinkhorn convergence and yield tighter clusters. Second, we plan to investigate the hard pruning of experts that exhibit consistently low utilization during training. Permanently removing these experts could reduce the model's memory footprint beyond just inference FLOPs, bridging the gap between MoEfication and model compression.

\section*{Acknowledgements}
We thank Aditi Raghunathan and Sankalp Dayal for valuable feedback on experimental design and ablation studies.

\section*{Impact Statement}
This paper presents work whose goal is to advance the field of Machine Learning by improving the inference efficiency of Large Language Models. By reducing the computational cost required for deployment, our method contributes to lowering the energy consumption and carbon footprint of foundation models. There are many potential societal consequences of our work, none of which we feel must be specifically highlighted here.

\newpage
\bibliography{main}
\bibliographystyle{icml2026}

\appendix
\clearpage
\section*{Appendix}
\section{The Assignment-Routing Gap}
\label{app:motivation_experiment}

The central claim of our method is that existing dense-to-MoE approaches optimize inadequate proxies for the FFN output, whereas DOT-MoE is output-aware. We provide two complementary pieces of evidence: a single-layer reconstruction analysis that isolates the assignment strategy under a fixed evaluation protocol, and a full-pipeline ablation that isolates it from training-recipe effects.

\subsection{Single-Layer Reconstruction}
\label{app:single_layer_mse}

We conduct a controlled single-layer analysis that directly measures reconstruction fidelity under different expert assignment strategies. For each method, we compute the mean squared error (MSE) between the output of the original dense FFN and its sparse MoE approximation, isolating the effect of expert construction and routing.

We analyze layer 31 of both LLaMA-2-7B and LLaMA-3-8B, partitioning the FFN intermediate dimension into experts of size $D = 128$ and applying top-$k = 10$ routing. Expert assignments are constructed using calibration data from the WikiText training split, and reconstruction error is evaluated on the WikiText-2 test set. We compare four assignment strategies: LLaMA-MoE v1, LLaMA-MoE v2, CMoE, and our proposed \name. All methods share identical expert granularity and routing configurations within each model.

Table~\ref{tab:motivation_mse_combined} shows a consistent performance gap across both LLaMA-2 and LLaMA-3. On LLaMA-2, LLaMA-MoE v1 incurs over $35\times$ higher reconstruction error than \name, LLaMA-MoE v2 nearly $9\times$, and CMoE more than $2\times$. The same pattern holds for LLaMA-3, where random and proxy-based assignments yield substantially higher error, while \name{} consistently achieves the lowest MSE. These results indicate that clustering neurons based on input-side statistics or intermediate activations is insufficient, and that preserving the FFN output requires explicitly modeling each neuron's interaction with the down-projection and residual stream.

\begin{table}[h!]
\centering
\caption{Single-layer reconstruction loss (MSE) on layer 31 for LLaMA-2 and LLaMA-3. All models use $D=128$ neurons per expert and top-$k=10$ routing. Calibration is performed on WikiText training data, and evaluation is conducted on WikiText-2. Lower is better.}
\label{tab:motivation_mse_combined}
\begin{tabular}{lcccc}
\toprule
\multirow{2}{*}{\textbf{Method}}
& \multicolumn{2}{c}{\textbf{LLaMA-2}}
& \multicolumn{2}{c}{\textbf{LLaMA-3}} \\
\cmidrule(lr){2-3} \cmidrule(lr){4-5}
& \textbf{Error} & \textbf{Rel.}
& \textbf{Error} & \textbf{Rel.} \\
\midrule
LLaMA-MoE v1  & 10.62 & +36x & 2.91 & +41.6x \\
LLaMA-MoE v2  & 2.92  & +9.9x & 0.43 & +6.1x \\
CMoE          & 0.61  & +2.1x & 0.15 & +2.1x \\
\rowcolor{blue!5} \textbf{DOT-MoE} & \textbf{0.29} & --- & \textbf{0.07} & --- \\
\bottomrule
\end{tabular}
\end{table}

\subsection{Output-Aware Assignment Ablation}
\label{app:output_aware_ablation}

To further isolate the contribution of the OT-based assignment from training-recipe effects, we run CMoE and DOT-MoE under an identical fine-tuning pipeline on Qwen2.5-7B at the same expert granularity ($E{=}8$, top-$k{=}2$), with the same data and training steps. Table~\ref{tab:output_aware_similarity} reports cosine similarity and MSE of hidden representations before and after the LM head against the dense teacher.

\begin{table}[h]
\centering
\caption{Representation similarity to the dense teacher on Qwen2.5-7B ($E{=}8$, top-$k{=}2$) under an identical fine-tuning pipeline. DOT-MoE preserves the dense residual-stream and logit geometry substantially better than CMoE.}
\label{tab:output_aware_similarity}
\setlength{\tabcolsep}{3pt}
\resizebox{\columnwidth}{!}{
\begin{tabular}{l cccc}
\toprule
\multirow{2}{*}{\textbf{Method}} & \textbf{Cos Sim} & \textbf{MSE} & \textbf{Cos Sim} & \textbf{MSE} \\
 & \textbf{(pre-LM)} & \textbf{(pre-LM)} & \textbf{(post-LM)} & \textbf{(post-LM)} \\
\midrule
CMoE & 0.55 & 21.1 & 0.49 & 16.6 \\
\rowcolor{blue!5} \textbf{DOT-MoE} & \textbf{0.84} & \textbf{8.2} & \textbf{0.71} & \textbf{10.5} \\
\bottomrule
\end{tabular}
}
\end{table}

Combined with the single-layer reconstruction analysis above and the training-dynamics comparison (Figure~\ref{fig:effect_of_init}), this confirms that the improvements come from the OT-based assignment itself and not from a stronger training recipe.

\section{Implementation Details}
\label{app:implementation_details}

\subsection{Hyperparameters}
Tables~\ref{tab:hyperparams_moefication} and~\ref{tab:hyperparams_sft} summarize the hyperparameters used for MoEfication and supervised fine-tuning, respectively. All hyperparameters are kept consistent across model families unless otherwise noted.

\begin{table}[h]
\centering
\caption{Hyperparameters for MoEfication.}
\label{tab:hyperparams_moefication}
\begin{tabular}{lll}
\toprule
\textbf{Hyperparameter} & \textbf{Symbol} & \textbf{Value} \\
\midrule
Expert size & $s$ & 128 \\
Sinkhorn iterations & $N$ & 50 \\
Sinkhorn temperature & $\tau$ & 0.1 \\
Learning rate & $\eta$ & $5 \times 10^{-4}$ \\
Weight decay & $\lambda$ & $10^{-4}$ \\
LR schedule & -- & Cosine \\
Warmup ratio & -- & 0.2 \\
Max gradient norm & -- & 1.0 \\
Batch size & -- & 64 \\
Sequence length & -- & 2048 \\
KL loss weight & $w_{\text{kl}}$ & 2.0 \\
CE loss weight & $w_{\text{ce}}$ & 1.0 \\
Z-loss weight & $w_z$ & $10^{-3}$ \\
Load balancing weight & $w_{\text{lb}}$ & $10^{-2}$ \\
\bottomrule
\end{tabular}
\end{table}

\begin{table}[h]
\centering
\caption{Hyperparameters for supervised fine-tuning.}
\label{tab:hyperparams_sft}
\begin{tabular}{ll}
\toprule
\textbf{Hyperparameter} & \textbf{Value} \\
\midrule
Learning rate & $5 \times 10^{-5}$ \\
Warmup ratio & 0.1 \\
LR schedule & Cosine \\
Batch size & 128 \\
Training tokens & 1.2B \\
Max gradient norm & 1.0 \\
Load balancing weight & $10^{-4}$ \\
\bottomrule
\end{tabular}
\end{table}

\subsection{Numerical Stability}

We implement several techniques to ensure numerical stability during training:
\subsubsection{Log-Domain Implementation.}
\label{app:log_domain_sinkhorn}
For numerical stability, we implement Sinkhorn iterations in log-space (Algorithm~\ref{alg:sinkhorn}). This avoids underflow when $\tau$ is small and enables stable computation even with thousands of neurons and experts.

\begin{algorithm}[]
\caption{Log-Domain Sinkhorn for Balanced Assignment}
\label{alg:sinkhorn}
\begin{algorithmic}[1]
\REQUIRE Assignment logits $\mathbf{A} \in \mathbb{R}^{d_{\text{ffn}}\times E}$, temperature $\tau$, iterations $N$
\ENSURE Soft Assignment Matrix $\mathbf{M}_{\text{soft}} \in [0,1]^{d_{\text{ffn}} \times E}$
\STATE $\mathbf{K} \leftarrow \mathbf{A} / \tau$
\STATE $\mathbf{u} \leftarrow \mathbf{0}, \quad \mathbf{v} \leftarrow \log(s) \cdot \mathbf{1}_E$
\FOR{$t = 1$ to $N$}
    \STATE $\mathbf{u} \leftarrow - \text{logsumexp}(\mathbf{K} + \mathbf{v}, \text{dim}=1)$
    \STATE $\mathbf{v} \leftarrow \log(s \cdot \mathbf{1}_E) - \text{logsumexp}(\mathbf{K} + \mathbf{u}, \text{dim}=0)$
\ENDFOR
\STATE $\mathbf{M}_{\text{soft}} \leftarrow \exp(\mathbf{K} + \mathbf{u} + \mathbf{v})$
\end{algorithmic}
\end{algorithm}

\subsubsection{Assignment Logits Precision.} The assignment logits $\mathbf{A}$ are maintained in FP32 regardless of model dtype to ensure stable Sinkhorn iterations. Router weights $\mathbf{W}_r$ use the model's native dtype.

\subsubsection{Temperature Annealing.} We linearly anneal the Sinkhorn temperature from $\tau_{\text{start}} = 1.0$ to $\tau_{\text{end}} = 0.1$ during the warmup phase. Higher temperatures early in training allow exploration of the assignment space; lower temperatures sharpen assignments as training progresses. Validation always uses $\tau_{\text{end}}$.

\section{Effect of Training Sparsity}
\label{app:sparsity_detailed}

Table~\ref{tab:sparsity_detailed} presents the complete per-benchmark results for the training sparsity ablation discussed in Section 4. We train two Qwen2.5-7B models at 50\% and 75\% FFN sparsity, then evaluate each across four inference sparsity levels.

\begin{table}[h]
\centering
\caption{Detailed benchmark performance under different training and inference sparsity configurations on Qwen2.5-7B. Models trained at higher sparsity (75\%) generalize better to extreme sparsity regimes at inference time, while both converge at low sparsity (30\%).}
\label{tab:sparsity_detailed}
\setlength{\tabcolsep}{4pt}
\resizebox{\columnwidth}{!}{
\begin{tabular}{c ccccc c}
\toprule
\textbf{Inference} & \textbf{ARC-e} & \textbf{ARC-c} & \textbf{PIQA} & \textbf{WinoG.} & \textbf{HellaS.} & \multirow{2}{*}{\textbf{Avg.}} \\
\textbf{Sparsity} & \scriptsize{[acc-n]} & \scriptsize{[acc-n]} & \scriptsize{[acc-n]} & \scriptsize{[acc]} & \scriptsize{[acc-n]} & \\
\midrule
\rowcolor{gray!10} \multicolumn{7}{c}{\textit{Trained at 50\% FFN Sparsity}} \\
90\% & 26.64 & 25.26 & 52.07 & 49.80 & 28.13 & 36.38 \\
75\% & 42.85 & 26.62 & 61.70 & 53.28 & 41.20 & 45.13 \\
50\% & 70.50 & 45.65 & 73.72 & 63.38 & 67.11 & 64.07 \\
30\% & 79.08 & 54.27 & 78.62 & 68.90 & 75.16 & 71.21 \\
\midrule
\rowcolor{gray!10} \multicolumn{7}{c}{\textit{Trained at 75\% FFN Sparsity}} \\
90\% & 34.81 & 21.93 & 56.04 & 50.28 & 32.56 & 39.12 \\
75\% & 59.97 & 35.24 & 68.82 & 59.98 & 54.21 & 55.64 \\
50\% & 72.77 & 48.55 & 76.22 & 66.77 & 70.45 & 66.95 \\
30\% & 80.35 & 53.92 & 79.27 & 69.53 & 75.83 & 71.78 \\
\bottomrule
\end{tabular}
}
\end{table}

\section{Training Objective}
\label{app:training_objective}

We jointly optimize the affinity matrix $\mathbf{A}$ and router weights $\mathbf{W}_{\text{r}}$ to minimize the discrepancy between dense and sparse outputs. Given a sequence of $T$ tokens, let $\mathbf{z}^{\text{dense}}_t, \mathbf{z}^{\text{MoE}}_t \in \mathbb{R}^{V}$ denote the output logits from the dense teacher and MoE student at position $t$, respectively.

\textbf{KL Divergence Loss.} We distill the dense model's output distribution into the MoE student. This loss directly minimizes the distribution gap between teacher and student outputs.

\textbf{Cross-Entropy Loss.} We additionally train on the standard language modeling objective. This ensures the MoE model maintains language modeling capability on the training distribution.

\textbf{Router Z-Loss.} Following~\citet{zoph2022st}, we penalize large router logits to improve numerical stability:
\begin{equation}
\mathcal{L}_z = \frac{1}{T}\sum_{t=1}^{T} \Bigl(\log\sum_{e=1}^E \exp(L_{t,e})\Bigr)^2
\label{eq:z_loss}
\end{equation}
where $L_{t,e}$ is the router logit (Equation~\ref{eq:router_logits}) for expert $e$ at token position $t$. This loss prevents the router from producing extremely large logits that can cause overflow in the softmax computation.

\textbf{Load Balancing Loss.} To prevent expert collapse and encourage uniform utilization, we include the auxiliary balancing loss from~\citet{shazeer2017sparsely}:
\begin{equation}
\mathcal{L}_{\text{bal}} = E \cdot \sum_{e=1}^E f_e \cdot \bar{p}_e
\label{eq:balance_loss}
\end{equation}
where $f_e = |\{i : e \in \mathcal{I}_i\}|/n$ is the fraction of tokens routed to expert $e$, and $\bar{p}_e = \frac{1}{n}\sum_{i=1}^{n} P_{i,e}$ is the average routing probability for expert $e$. The product $f_e \cdot \bar{p}_e$ is minimized when experts are utilized uniformly.

\textbf{Total Objective.} The complete training loss combines all components:
\begin{equation}
\mathcal{L} = w_{\text{kl}}\mathcal{L}_{\text{KL}} + w_{\text{ce}}\mathcal{L}_{\text{CE}} + w_z\mathcal{L}_z + w_{\text{bal}}\mathcal{L}_{\text{bal}}
\label{eq:total_loss}
\end{equation}
where $w_{\text{kl}}, w_{\text{ce}}, w_z, w_{\text{bal}}$ are weighting hyperparameters (see Table~\ref{tab:hyperparams_moefication}). 

\section{Sparse MoE Computation}
\label{app:sparse_moe_computation}

During the alignment phase, we simulate sparse MoE computation without materializing separate expert weights. Given the neuron assignment $\mathbf{M}$ and routing mask $\mathbf{R}$, we compute the sparse MoE output by masking the intermediate activations:
\begin{equation}
\hat{\mathbf{Y}} = \bigl(\mathbf{H} \odot (\mathbf{R}\mathbf{M}^{\top})\bigr) \mathbf{W}_{\text{down}}
\label{eq:moe_output}
\end{equation}
where $\mathbf{H} \in \mathbb{R}^{n \times d_{\text{ffn}}}$ is the intermediate activation from Eq.~\ref{eq:ffn_intermediate} and $\odot$ denotes element-wise multiplication.

The matrix product $\mathbf{R}\mathbf{M}^\top \in \{0,1\}^{n \times d_{\text{ffn}}}$ composes two levels of selection:
\begin{itemize}[noitemsep]
    \item $\mathbf{R} \in \{0,1\}^{n \times E}$: which experts are active for each token
    \item $\mathbf{M}^\top \in \{0,1\}^{E \times d_{\text{ffn}}}$: which neurons belong to each expert
    \item $\mathbf{R}\mathbf{M}^\top \in \{0,1\}^{n \times d_{\text{ffn}}}$: which neurons are active for each token
\end{itemize}
Since each token activates $k$ experts and each expert contains $s$ neurons, only $k \cdot s$ out of $d_{\text{ffn}}$ neurons contribute to each token's output. This masking-based formulation enables efficient training on the original dense weights while simulating sparse computation.

After alignment training converges, we extract the final binary assignment $\mathbf{M}$ and use it to partition the dense FFN weights into $E$ separate expert modules. For each expert $e$, we slice the corresponding rows from $\mathbf{W}_{\text{gate}}$, $\mathbf{W}_{\text{up}}$, and columns from $\mathbf{W}_{\text{down}}$ according to the neurons in cluster $\mathcal{C}_e = \{i : M_{i,e} = 1\}$. The resulting model is a standard MoE architecture compatible with existing sparse inference frameworks.

\section{Expert Specialization and Utilization}
\begin{figure}[ht]
    \centering
    \includegraphics[width=0.7\linewidth]{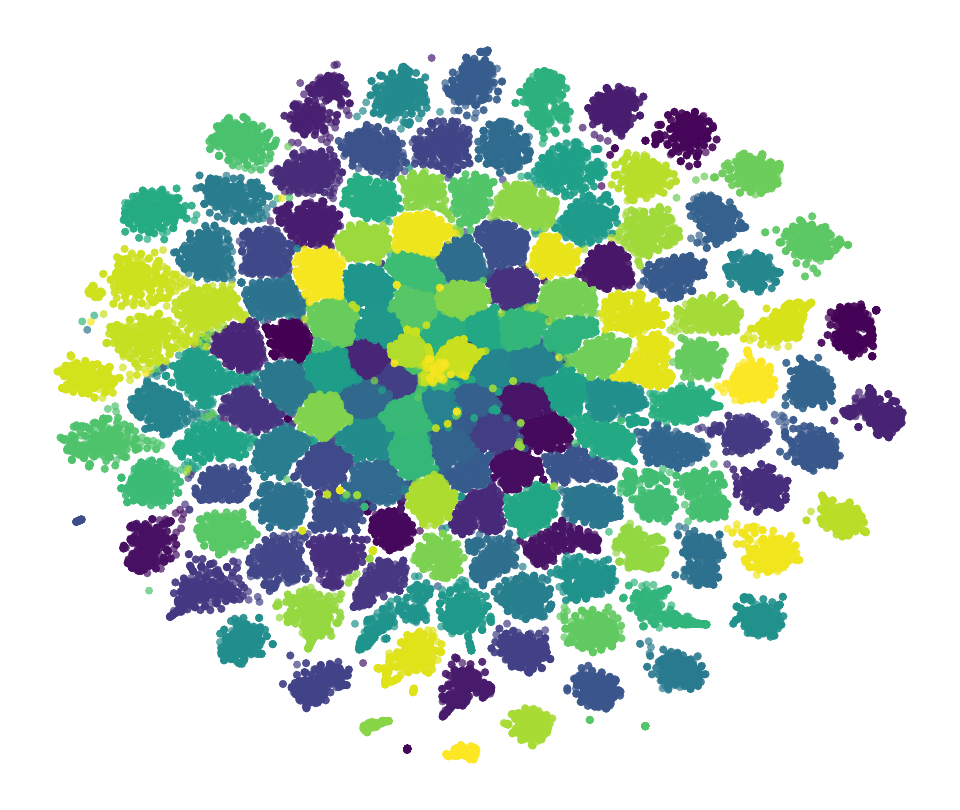}
    \caption{t-SNE visualization of expert output activations at layer 9 for Qwen2.5-7B. Each color represents a different expert. The clear clustering indicates that experts learn distinct, well-separated representations.}
    \label{fig:tsne_appendix}
\end{figure}

\textbf{Expert Specialization. }To understand the learned behavior of our converted MoE model, we visualize expert output activations using t-SNE \cite{maaten2008visualizing}. We collect activation vectors from the expert outputs across a diverse set of input samples and project them into two dimensions. Figure~\ref{fig:tsne_appendix} shows the resulting visualization for layer 9, where each color represents a different expert. The visualization reveals clear clustering structure, indicating that experts learn to specialize in processing distinct types of inputs. Activations from the same expert tend to cluster together in the embedding space, forming well-separated regions. The clear separation between expert clusters suggests that our method successfully partitions the representation space, with minimal redundancy across experts.

\begin{figure}[ht]
\includegraphics[width=\linewidth]{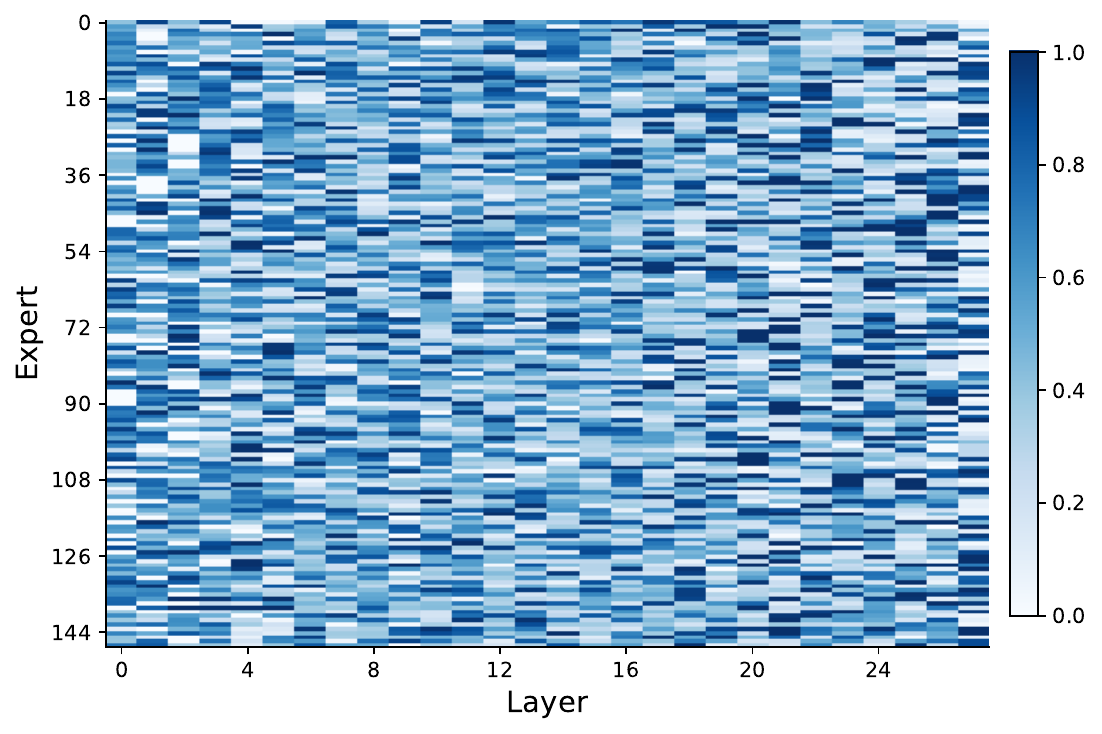}
    \caption{Expert token allocation across transformer layers for Qwen2.5-7B with 50\% sparsity on the WikiText-2 dataset.}
\label{fig:expert_utilization}
\end{figure}

\textbf{Expert Utilization. }To analyze routing behavior after MoEfication, we collect expert token allocation statistics across all transformer layers on the WikiText-2 dataset for Qwen2.5-7B with 50\% sparsity. Figure \ref{fig:expert_utilization} visualizes the proportion of tokens routed to each expert at every layer. Overall, expert routing remains well balanced across most layers, with no evidence of severe expert collapse.

\begin{table*}[h]
\centering
\caption{Comparison with LTE on LLaMA-2-7B ($E{=}86$) at matched FFN sparsity. DOT-MoE outperforms LTE by +6.0 points on average while keeping per-token compute constant through top-$k$ routing.}
\label{tab:lte_comparison}
\begin{tabular}{lc cccccc c}
\toprule
\textbf{Method} & \textbf{Active FFN} & \textbf{BoolQ} & \textbf{SciQ} & \textbf{PIQA} & \textbf{WinoG.} & \textbf{ARC-C} & \textbf{HellaS.} & \textbf{Avg.} \\
\midrule
LTE ($\eta{=}3.0$) & 29\% & 68.9 & 82.5 & 59.5 & 62.7 & 33.9 & 56.2 & 60.6 \\
\rowcolor{blue!5} \textbf{DOT-MoE} & \textbf{25\%} & \textbf{72.5} & \textbf{94.3} & \textbf{69.3} & \textbf{62.5} & \textbf{40.9} & \textbf{60.2} & \textbf{66.6} \\
\bottomrule
\end{tabular}
\end{table*}

\section{Extension to Attention Layers}
\label{sec:attention_moe}

\begin{table*}[ht]
\centering
\caption{Attention MoEfication on Qwen2.5-7B at 50\% attention sparsity. OT-based head assignment outperforms random assignment by +17.9 points on average.}
\label{tab:attention_moe_results}
\begin{tabular}{l cccccc c}
\toprule
\textbf{Method} & \textbf{BoolQ} & \textbf{SciQ} & \textbf{PIQA} & \textbf{WinoG.} & \textbf{ARC-C} & \textbf{HellaS.} & \textbf{Avg.} \\
\midrule
Random assign. + router & 45.1 & 60.5 & 56.8 & 51.8 & 27.7 & 35.2 & 46.2 \\
\rowcolor{blue!5} \textbf{DOT-MoE (Attn)} & \textbf{60.1} & \textbf{87.2} & \textbf{74.2} & \textbf{56.0} & \textbf{44.4} & \textbf{62.6} & \textbf{64.1} \\
\bottomrule
\end{tabular}
\end{table*}

DOT-MoE extends to multi-head attention by treating heads as the units to be grouped into experts, mirroring the FFN setting where neurons are grouped. Given $N_h$ query heads of dimension $d_h$, we form $E_{\text{attn}} = N_h / s_h$ experts of $s_h$ heads and activate $k_{\text{attn}}$ per token via a separate router $\mathbf{W}_{r}^{\text{attn}}$. A learnable affinity matrix $\mathbf{A}_{\text{attn}} \in \mathbb{R}^{N_h \times E_{\text{attn}}}$ with marginals $(\mathbf{1}_{N_h},\, s_h \mathbf{1}_{E_{\text{attn}}})$ is optimized via the same log-domain Sinkhorn iterations and discretized through the same STE as Eq.~\ref{eq:assignment_ste}. During training, all heads are computed with frozen dense weights; sparsity is realized by masking the concatenated head outputs before the output projection $\mathbf{W}_O$, yielding an expression identical in form to the FFN reconstruction (Eq.~\ref{eq:moe_output}). For Grouped Query Attention~\citep{ainslie2023gqa}, assignment operates at query-head granularity and all KV heads are computed unconditionally; sparsity therefore lives in the Q and O projections, which dominate attention parameters.

To validate the formulation, we evaluate attention-only DOT-MoE on Qwen2.5-7B at 50\% attention sparsity ($N_h{=}28$, $E_{\text{attn}}{=}14$, $s_h{=}2$, $k_{\text{attn}}{=}7$) against a random head-assignment baseline with a trained router under the same training recipe. Table~\ref{tab:attention_moe_results} shows that OT-based assignment outperforms the baseline by {+17.9 points} on average (64.1 vs.\ 46.2). Two points are worth noting. First, the search space is far smaller than for the FFN (28 heads vs.\ 18{,}944 neurons), so attention assignment converges faster and with less gradient variance; the attention router itself matches the FFN router architecturally and adds negligible compute. Second, because attention parameters are roughly one third of FFN parameters, attention-only MoEfication yields modest overall compression; combining it with FFN MoEfication (joint MLP+Attention) is a direct extension.

\begin{table*}[h]
\centering
\caption{Structural settings of dense-to-MoE conversion methods. Our primary baselines match DOT-MoE's problem setting: parameter-preserving, activation-agnostic, fixed-active-params per token, and softmax top-$k$ routing compatible with standard MoE serving frameworks.}
\label{tab:method_positioning}
\begin{tabular}{l cccccc}
\toprule
\multirow{2}{*}{\textbf{Method}} & \textbf{Param.} & \textbf{Act.} & \textbf{Fixed} & \multirow{2}{*}{\textbf{Routing}} & \textbf{Expert} & \textbf{MoE} \\
 & \textbf{Preserv.} & \textbf{Agnostic} & \textbf{Compute} & & \textbf{Assignment} & \textbf{Serve} \\
\midrule
MoEfication     & \checkmark & $\times$ (ReLU) & \checkmark & Top-$k$ & Weight clustering & \checkmark \\
DejaVu          & \checkmark & $\times$ (ReLU) & $\times$ & Per-input pred. & Activation pred. & $\times$ \\
LLaMA-MoE       & \checkmark & \checkmark & \checkmark & Top-$k$ & Random partition & \checkmark \\
LLaMA-MoE-v2    & \checkmark & \checkmark & \checkmark & Top-$k$ & Act./grad.\ heur. & \checkmark \\
CMoE            & \checkmark & \checkmark & \checkmark & Top-$k$ & Co-activation clust. & \checkmark \\
LTE             & \checkmark & \checkmark & $\times$ & Sigmoid+thresh. & $k$-means + router & $\times$ \\
Read-ME         & $\times$ (2.4$\times$) & \checkmark & \checkmark & Pre-gating & Activation-based & \checkmark \\
\midrule
\rowcolor{blue!5} \textbf{DOT-MoE} & \textbf{\checkmark} & \textbf{\checkmark} & \textbf{\checkmark} & \textbf{Top-$k$} & \textbf{Learned (OT)} & \textbf{\checkmark} \\
\bottomrule
\end{tabular}
\end{table*}

\section{Training Overhead}
\label{app:training_overhead}

We profile DOT-MoE's alignment phase on 8$\times$H100 to quantify the cost of Sinkhorn iterations and straight-through estimation relative to a standard dense forward/backward pass. Sinkhorn iterations account for only $\sim 2\%$ of the total forward-and-backward time. All DOT-MoE-specific operations combined add $\sim 15\%$ overhead per training step over a standard dense training step; most of this overhead comes from hard-assignment matrix construction rather than from Sinkhorn itself. We currently run the greedy rounding on CPU and incur CPU-to-GPU transfer overhead; a dedicated GPU kernel (e.g., using a parallel bucketing or priority-queue primitive) would remove most of this cost. Crucially, this overhead is incurred \emph{only during the alignment phase}: once alignment converges, the extracted MoE model is a standard fused-expert architecture with no Sinkhorn or STE at inference time.

\section{Additional Dense-to-MoE Baselines}
\label{app:additional_baselines}

\subsection{Comparison with LTE}
\label{app:lte_comparison}

LTE~\citep{2024Haizhonglte} controls sparsity through a scalar $\eta$ and activates a variable number of experts per token via sigmoid thresholding. To enable a fair comparison, we evaluate DOT-MoE at matching FFN sparsity on LLaMA-2-7B with $E{=}86$ experts (Table~\ref{tab:lte_comparison}).

At \emph{lower} FFN sparsity (25\% vs.\ 29\%), DOT-MoE still outperforms LTE by +6.0 points on average. Beyond accuracy, LTE's sigmoid routing activates a variable number of experts per token, leading to unpredictable per-token compute; DOT-MoE uses softmax top-$k$ routing, which keeps compute per token constant and is compatible with standard fused-MoE serving kernels.

\begin{table*}[h]
\centering
\caption{Controlled same-granularity comparison ($E{=}8$, top-$k{=}2$, 1.2B fine-tuning tokens). At CMoE's own granularity, DOT-MoE consistently outperforms CMoE across all three architectures.}
\label{tab:same_granularity}
\begin{tabular}{ll cccccc c}
\toprule
\textbf{Model} & \textbf{Method} & \textbf{BoolQ} & \textbf{SciQ} & \textbf{PIQA} & \textbf{WinoG.} & \textbf{ARC-C} & \textbf{HellaS.} & \textbf{Avg.} \\
\midrule
\multirow{2}{*}{Qwen2.5-7B} & CMoE & 69.4 & 93.8 & 67.6 & 59.0 & 37.8 & 57.2 & 64.1 \\
 & \cellcolor{blue!5}\textbf{DOT-MoE} & \cellcolor{blue!5}\textbf{75.3} & \cellcolor{blue!5}\textbf{94.6} & \cellcolor{blue!5}\textbf{69.3} & \cellcolor{blue!5}\textbf{63.7} & \cellcolor{blue!5}\textbf{45.8} & \cellcolor{blue!5}\textbf{57.6} & \cellcolor{blue!5}\textbf{67.7} \\
\midrule
\multirow{2}{*}{LLaMA-2-7B} & CMoE & 55.0 & 77.5 & 57.1 & 54.1 & 27.6 & 38.8 & 51.7 \\
 & \cellcolor{blue!5}\textbf{DOT-MoE} & \cellcolor{blue!5}\textbf{73.0} & \cellcolor{blue!5}\textbf{93.2} & \cellcolor{blue!5}\textbf{70.5} & \cellcolor{blue!5}\textbf{61.3} & \cellcolor{blue!5}\textbf{40.2} & \cellcolor{blue!5}\textbf{58.2} & \cellcolor{blue!5}\textbf{66.0} \\
\midrule
\multirow{2}{*}{LLaMA-3-8B} & CMoE & 71.1 & 94.4 & 69.5 & 59.5 & 38.2 & 55.3 & 64.7 \\
 & \cellcolor{blue!5}\textbf{DOT-MoE} & \cellcolor{blue!5}\textbf{73.8} & \cellcolor{blue!5}\textbf{94.6} & \cellcolor{blue!5}\textbf{70.5} & \cellcolor{blue!5}\textbf{62.7} & \cellcolor{blue!5}\textbf{40.4} & \cellcolor{blue!5}\textbf{60.3} & \cellcolor{blue!5}\textbf{67.0} \\
\bottomrule
\end{tabular}
\end{table*}

\subsection{Positioning Among Dense-to-MoE Methods}
\label{app:method_positioning}

To clarify why our primary baselines are CMoE and LLaMA-MoE(-v2), we compare the structural settings of dense-to-MoE methods in Table~\ref{tab:method_positioning}. Methods that use ReLU-specific activation patterns (MoEfication, DejaVu), inflate total parameters (Read-ME), or use variable-compute routing (DejaVu, LTE) address a different problem setting than ours. Our baselines share the parameter-preserving, fixed-active-params, softmax top-$k$ setting.

Read-ME is excluded as a primary baseline because it inflates parameters by $2.4\times$ (7B dense to 17B MoE), placing it in the upcycling rather than parameter-preserving category. MoEfication and DejaVu are excluded because they were designed for ReLU-based encoder architectures and do not transfer to SwiGLU decoder LLMs.

\section{Controlled Same-Granularity Comparison}
\label{app:same_granularity}

A natural concern in dense-to-MoE comparisons is that different expert granularities can confound the contribution of the assignment strategy. To address this, we run DOT-MoE at CMoE's own granularity ($E{=}8$, top-$k{=}2$) with 1.2B fine-tuning tokens across three model families. Table~\ref{tab:same_granularity} reports the controlled comparison.

At CMoE's own granularity, DOT-MoE outperforms CMoE by {+3.6} on Qwen2.5-7B, {+14.3} on LLaMA-2-7B, and {+2.3} on LLaMA-3-8B. Moreover, DOT-MoE at $E{=}8$ nearly matches DOT-MoE at $E{=}148$ on the same active-parameter budget (e.g., 67.7 vs.\ 67.2 on Qwen2.5-7B), indicating that DOT-MoE's advantage is not an artifact of finer expert granularity. We also verified that CMoE at DOT-MoE's default granularity ($E{=}148$) achieves 64.1 avg on Qwen2.5-7B versus DOT-MoE's 67.2, so the conclusion holds in both directions.

\section{Scalability}
\label{app:scalability}

\subsection{Scaling to 32B Parameters}
\label{app:qwen32b}

DOT-MoE's alignment phase only trains assignment logits and router weights (under 2\% of model parameters), so the method itself is not the bottleneck at scale; the frozen dense model's forward pass dominates cost. To verify scalability, we evaluate DOT-MoE on Qwen2.5-32B at 25\% active parameters (Table~\ref{tab:qwen32b_scalability}).

\begin{table}[h]
\centering
\caption{Scalability to Qwen2.5-32B at 25\% active parameters. Zero-shot performance on common-sense reasoning benchmarks.}
\label{tab:qwen32b_scalability}
\setlength{\tabcolsep}{3pt}
\resizebox{\columnwidth}{!}{
\begin{tabular}{l cccccc c}
\toprule
\textbf{Method} & \textbf{BoolQ} & \textbf{SciQ} & \textbf{PIQA} & \textbf{WinoG.} & \textbf{ARC-C} & \textbf{HellaS.} & \textbf{Avg.} \\
\midrule
CMoE & 43.5 & 34.1 & 52.8 & 49.6 & 23.7 & 28.8 & 38.8 \\
\rowcolor{blue!5} \textbf{DOT-MoE} & \textbf{81.0} & \textbf{95.4} & \textbf{74.8} & \textbf{67.6} & \textbf{49.1} & \textbf{70.6} & \textbf{73.1} \\
\bottomrule
\end{tabular}
}
\end{table}

At 32B parameters, DOT-MoE improves the benchmark average by {+34.3 points} (73.1 vs.\ 38.8) over CMoE, confirming that the OT-based assignment holds up as model scale increases.

\subsection{Robustness to Sequence Length}
\label{app:seqlen}

The neuron-to-expert assignment and router operate per token and are independent of sequence length, so DOT-MoE applies directly to longer contexts. We evaluate WikiText-2 word perplexity on Qwen2.5-7B (with 1.2B fine-tuning tokens) using rolling log-likelihood with varying maximum context windows. The document-level WikiText\footnote{EleutherAI/wikitext\_document\_level} split is used; documents exceeding the context window are split into rolling windows with sliding overlap.

\begin{table}[h]
\centering
\caption{WikiText-2 word perplexity on Qwen2.5-7B (with 1.2B fine-tuning tokens) at different maximum sequence lengths. DOT-MoE's advantage over CMoE is consistent across context windows from 2K to 32K.}
\label{tab:seqlen_ppl}
\setlength{\tabcolsep}{6pt}
\begin{tabular}{l cccc}
\toprule
\textbf{Method} & \textbf{2048} & \textbf{4096} & \textbf{8192} & \textbf{32768} \\
\midrule
Dense (Qwen2.5-7B) & 9.50 & 9.03 & 8.78 & 8.73 \\
\midrule
CMoE (E8S1A1) & 18.83 & 17.62 & 16.90 & 16.84 \\
\rowcolor{blue!5} \textbf{DOT-MoE (E8A2)} & \textbf{16.47} & \textbf{15.48} & \textbf{14.91} & \textbf{14.84} \\
\bottomrule
\end{tabular}
\end{table}

DOT-MoE maintains a consistent $\sim 2$ PPL improvement over CMoE across all context lengths up to 32K tokens, confirming that per-token routing is robust to long contexts.
\end{document}